\DocumentMetadata{}
\documentclass[sigconf]{acmart}

\usepackage{hyperxmp}  
\usepackage{algorithm}
\usepackage{algorithmic}
\usepackage{tabularx} 
\usepackage{booktabs} 
\usepackage{subcaption}

\AtBeginDocument{%
  \providecommand\BibTeX{{%
    \normalfont B\kern-0.5em{\scshape i\kern-0.25em b}\kern-0.8em\TeX}}}

\copyrightyear{2024}
\acmYear{2024}
\setcopyright{acmlicensed}\acmConference[GECCO '24]{Genetic and Evolutionary Computation Conference}{July 14--18, 2024}{Melbourne, VIC, Australia}
\acmBooktitle{Genetic and Evolutionary Computation Conference (GECCO '24), July 14--18, 2024, Melbourne, VIC, Australia}
\acmDOI{10.1145/3638529.3654064}
\acmISBN{979-8-4007-0494-9/24/07}

\begin{document}

\title{Efficient Multi-Objective Neural Architecture Search via Pareto Dominance-based Novelty Search}

\author{An Vo}
\affiliation{%
  \institution{University of Information Technology}
  \city{Ho Chi Minh City}
  \country{Vietnam}}
  \affiliation{%
  \institution{Vietnam National University}
  \city{Ho Chi Minh City}
  \country{Vietnam}}
\email{19520007@gm.uit.edu.vn}

\author{Ngoc Hoang Luong}
\authornote{Corresponding author.}
\affiliation{%
  \institution{University of Information Technology}
  \city{Ho Chi Minh City}
  \country{Vietnam}}
  \affiliation{%
  \institution{Vietnam National University}
  \city{Ho Chi Minh City}
  \country{Vietnam}}
\email{hoangln@uit.edu.vn}

\renewcommand{\shortauthors}{An Vo and Ngoc Hoang Luong}

\begin{abstract}
Neural Architecture Search (NAS) aims to automate the discovery of high-performing deep neural network architectures. Traditional objective-based NAS approaches typically optimize a certain performance metric (e.g., prediction accuracy), overlooking large parts of the architecture search space that potentially contain interesting network configurations. Furthermore, objective-driven population-based metaheuristics in complex search spaces often quickly exhaust population diversity and succumb to premature convergence to local optima. This issue becomes more complicated in NAS when performance objectives do not fully align with the actual performance of the candidate architectures, as is often the case with training-free metrics. While training-free metrics have gained popularity for their rapid performance estimation of candidate architectures without incurring computation-heavy network training, their effective incorporation into NAS remains a challenge. This paper presents the Pareto Dominance-based Novelty Search for multi-objective NAS with Multiple Training-Free metrics (MTF-PDNS). Unlike conventional NAS methods that optimize explicit objectives, MTF-PDNS promotes population diversity by utilizing a novelty score calculated based on multiple training-free performance and complexity metrics, thereby yielding a broader exploration of the search space. Experimental results on standard NAS benchmark suites demonstrate that MTF-PDNS outperforms conventional methods driven by explicit objectives in terms of convergence speed, diversity maintenance, architecture transferability, and computational costs.
\end{abstract}




\maketitle

\section{Introduction}
Neural Architecture Search (NAS), a subfield of automated machine learning, aims to efficiently automate the design process of neural network architectures~\cite{nas_survey,nas_1000_papers}. It is driven by the goal of discovering optimal configurations in a large-scale design space with minimal manual intervention. Vanilla NAS methods~\cite{nas_rl,amoebanet} are often designed to optimize specific performance metrics, and these methods tend to prioritize optimization of performance objectives. 
However, this focus can inadvertently constrain the exploration of the vast search space, overlooking potentially innovative configurations that lie outside the objective-driven search trajectories~\cite{why_greeatness_book}. 


A fundamental challenge in NAS is that the evaluations of numerous candidate architectures are prohibitively time-consuming~\cite{nas_rl,amoebanet}. To address this, many performance predictors~\cite{performance_predictors} have been proposed recently, notably the emergence of training-free metrics~\cite{zero_cost}. While these metrics can provide a quick and cost-effective way to estimate the performance of different architectures, their use is not without challenges. One of the main arguments is that they may not always correlate well with the final test accuracy~\cite{zero_cost}. Consequently, over-reliance on these metrics as the direct objective in NAS, due to their insufficient informativeness about the performance of candidate architectures, could potentially lead to bias and deception in the search process~\cite{nas_bench_suite_zero,eval_performance_estimartos}. These inaccurate performance objectives can misguide the search process, leading the search to converge prematurely to bad local optima - architectures that seem effective according to performance predictors but are not the best solutions in terms of actual performance. Thus, the search process can neglect the exploration of other potentially high-performing architectures that could exist elsewhere in the search space~\cite{ns_abonding_obj,ns_emperical_study,ns_global_opt}.


Novelty Search (NS) has been proposed as an alternative approach that rewards divergence in characteristic features or behaviors of solutions from those previously explored in the search process, rather than the optimization of performance objectives alone~\cite{why_greeatness_book,ns_abonding_obj,ns_theoretical_perspective}. NS has shown potential in various domain~\cite{ns_robotics,ns_game_content}, including NAS~\cite{ns_enas_gecco,ns_nas_one_shot_sampling,nsas_nas_cvpr}. In NAS, the novelty score is calculated based on the distinctive descriptors of each architecture in relation to others in the current population, and possibly as well as explored architectures in the previous generations stored in an archive. When a novelty score is pursued, the search process is encouraged to perform more exploration, thereby obtaining diverse solutions, potentially from simple to increasingly more complex ones. The novelty method can lead to the discovery of a wider variety of neural architectures, some of which may be more effective, that could be overlooked by conventional search methods. This shift in focus from rigidly pursuing performance objectives, which can lead to convergence in a single region of search space, to promoting exploration and novelty, can help avoid deception and premature convergence~\cite{ns_emperical_study,ns_abonding_obj,ns_global_opt}.


In this study, we propose the \textbf{P}areto \textbf{D}ominance-based \textbf{N}ovelty \textbf{S}earch for NAS with \textbf{M}ultiple \textbf{T}raining-\textbf{F}ree metrics (MTF-PDNS), which introduces a novelty score that utilizes training-free performance and complexity metrics to enhance evolutionary multi-objective NAS~\cite{nsga-net-v1,nsga-net-v2}. This score is designed to integrate both novelty and fitness of architectures, guiding a Genetic Algorithm (GA) as the primary objective. 
To complement this, we maintain an elitist archive~\cite{elitist_archive}, which preserves training-free performance and complexity metrics of non-dominated architectures. Our method only utilizes archives to calculate novelty scores without directly optimizing the metrics.
Moreover, we take into consideration various aspects of neural network architectures, prioritizing not just performance but also complexity, which makes it more applicable to real-world scenarios. To validate our approach, we conduct extensive experiments across widely-used NAS benchmarks, including NAS-Bench-101~\cite{nas_bench_101}, NAS-Bench-201~\cite{nas_bench_201}, and NAS-Bench-1Shot1~\cite{nas_bench_1shot1}. The experimental results underscore the superiority of our method in the context of Multi-Objective NAS (MONAS), indicating its ability to effectively navigate the search space and efficiently explore a wide range of neural network architectures.

\section{Preliminaries}\label{sec:preliminaries}
\subsection{Multi-objective NAS}
Multi-Objective NAS (MONAS~\cite{nsga-net-v1,nsga-net-v2}) ideally aims to find within the architecture search space $\Omega$ a Pareto set $\mathcal{P}_S\subset\Omega$ of neural network topologies that exhibit optimal trade-offs regarding $m$ objectives.
These objectives often compete with each other, such as minimizing error while minimizing computational resource usage, model complexity, or inference latency. Assuming minimization for all objectives, the MONAS problem can be defined as follows:
\begin{equation}
\min \mathbf{F}(\mathbf{x}) = [f_1(\mathbf{x}), f_2(\mathbf{x}), \ldots, f_m(\mathbf{x})]: \Omega \rightarrow \mathbb{R}^m,
\end{equation}
where $\Omega$ is the search space and $f_i(\mathbf{x})$ denotes the $i$-th objective function to be minimized. Let $\mathbf{x}, \mathbf{y} \in \Omega$ be two architectures. We say that $\mathbf{x}$ Pareto dominates $\mathbf{y}$ (denoted as $\mathbf{x} \prec \mathbf{y}$) if:

\begin{equation}
\begin{cases}
\forall i \in \{1, \ldots, m\}, f_i(\mathbf{x}) \leq f_i(\mathbf{y}) \\
\exists j \in \{1, \ldots, m\}, f_j(\mathbf{x}) < f_j(\mathbf{y})
\end{cases}
\end{equation}

The first condition says that architecture $\mathbf{x}$ is no worse than architecture $\mathbf{y}$ in all objectives, and the second condition says that architecture $\mathbf{x}$ is strictly better than architecture $\mathbf{y}$ in at least one objective. The Pareto set $\mathcal{P}_S$ of an MONAS problem thus consists of all Pareto-optimal architectures $\mathbf{x}\in\Omega$ such that there is no other architecture $\mathbf{y}\in\Omega$ dominating $\mathbf{x}$:
$
\mathcal{P}_S = \{\mathbf{x} \in \Omega \mid \nexists \mathbf{y} \in \Omega,  \mathbf{y} \prec \mathbf{x}\}
$.
The Pareto front $\mathcal{P}_F$ is defined as the set of vectors of objective values of architectures in $\mathcal{P}_S$, representing all the optimal trade-offs among the objectives:
$
    \mathcal{P}_F=\mathbf{F}(\mathcal{P}_S)=\{\mathbf{F}(\mathbf{x}) ~\vert~ \mathbf{x}\in\mathcal{P}_S\}
$.
Instead of finding the entire Pareto set $\mathcal{P}_S$, in real-world MONAS, it is often practical to find an approximation set $A$ containing candidate architectures that do not Pareto dominate each other and $\mathbf{F}(A)$ closely approximates $\mathcal{P}_F$ in the objective space.

\begin{figure*}[ht!]
    \centering
    \includegraphics[width=0.95\textwidth]{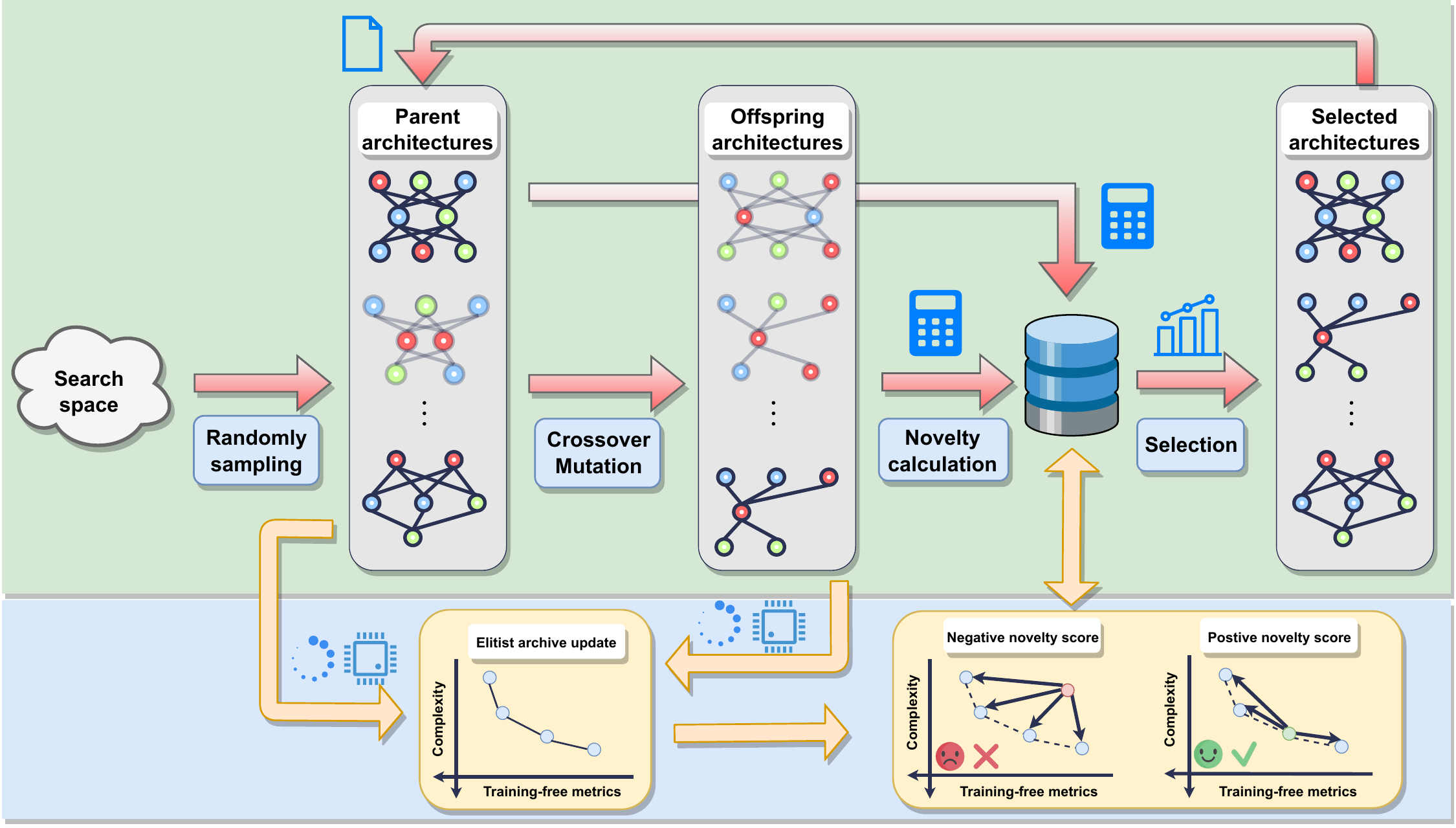}
    \caption{Illustration of Pareto Dominance-based Novelty Search. In the elitist archive update (yellow background), if we use descriptors derived from training-based metrics to calculate novelty scores, the x-axis represents validation error. In contrast, if we use descriptors derived from training-free metrics, the x-axis represents training-free metrics as shown in the figure.}
    \label{fig:PDNS}
\end{figure*}

\subsection{Training-free metrics for NAS}\label{sec:tf_metrics}
Training-free metrics~\cite{zero_cost} are used to estimate the quality of neural network architectures without having to firstly train them to obtain proper network weight values. 
These metrics are useful in NAS as they dramatically reduce the computational cost associated with evaluating candidate architectures. 
In this section, we discuss three such metrics: Synaptic Flow (\texttt{synflow})~\cite{synflow}, Jacobian Covariance (\texttt{jacov})~\cite{jacob_cov}, and \texttt{snip}~\cite{snip}. Both \texttt{synflow} and \texttt{snip} are originally proposed as metrics for network pruning, which is the process of removing less important weights from a neural network in order to reduce its complexity. However, they differ in how they determine the importance of weights. The intuitions and calculations for these training-free metrics are provided in the supplementary material.

Krishnakumar et al.~\cite{nas_bench_suite_zero} analyze the biases inherent in training-free metrics on NAS benchmarks and indicate that training-free metrics exhibit varying degrees of biases. For instance, \texttt{synflow} was found to have a strong bias towards larger cell sizes on NAS-Bench-201, meaning its preference for larger architectures. Similarly, both \texttt{snip} and \texttt{synflow} demonstrate a high bias with respect to the number of architecture parameters. Recognizing these biases is crucial for understanding the limitations of training-free metrics and also for developing methods to reduce bias, which can lead to improved performance in NAS~\cite{nas_bench_suite_zero}.

Many studies have been conducted recently to effectively employ training-free metrics for NAS. 
For example, FreeGEA~\cite{free_gea} combines \texttt{synflow}, linear regions, and the number of skip layers as a single objective to be optimized. In contrast, E-TF-MOENAS~\cite{infor_sci} optimizes \texttt{synflow}, \texttt{jacov}, and the number of parameters (or FLOPs) simultaneously as separate objectives. Abdelfattah et al.~\cite{zero_cost} propose a voting mechanism utilizing all three metrics \texttt{synflow}, \texttt{jacov}, and \texttt{snip}. The voting results are shown to achieve a strong correlation with network accuracy on various NAS benchmarks. 
Using multiple training-free metrics is a suggested way to increase NAS performance, while mitigating the bias of each training-free metric~\cite{nas_bench_suite_zero}.
In this work, our proposed novelty search approach also utilizes \texttt{synflow}, \texttt{jacov}, and \texttt{snip} simultaneously. However, these metrics are used as character descriptors for calculating the novelty score of an architecture, rather than being directly optimized as performance objectives as in aforementioned approaches.

\subsection{Novelty Search}\label{sec:ns}

Suppose we have a set of individuals $P = \{\mathbf{a}_1, \mathbf{a}_2, \ldots, \mathbf{a}_N\}$, where each individual $\mathbf{a}$ has a set of descriptors $\phi(\mathbf{a})$. The novelty score $\eta(\mathbf{x})$ for an individual $\mathbf{x}$ can be calculated as follows:

\begin{equation}
    \eta(\mathbf{x}) = \frac{1}{N} \sum_{i=1}^N \delta(\phi(\mathbf{x}), \phi(\mathbf{a}_i))
\end{equation}

In this formula, $\delta$ represents a distance function (e.g., Euclidean distance), which quantifies the difference between the descriptor set $\phi(\mathbf{x})$ of individual $\mathbf{x}$ and those of each other individual $\mathbf{a}_i$ in the set of individuals $P$. This aims to encourage the algorithm's exploration of less-visited parts of the descriptor space, rewarding individuals with unique descriptors. In NAS, this approach enhances the likelihood of identifying architectures that may be overlooked when the exploration of the search space is driven directly by performance metrics. However, designing a novelty score $\eta$ requires task-specific knowledge as we must find a way to strike a balance between the novelty and the quality of architectures. 

Despite the potential, the application of NS in NAS has been explored in a relatively limited number of studies. Zhang et al.~\cite{ns_nas_one_shot_sampling} proposed a mechanism based on NS to apply an architecture sampling controller in one-shot NAS~\cite{one_shot_nas_1,one_shot_nas_2}. The work of Sinha et al.~\cite{ns_enas_gecco}, on the other hand, utilized NS for both supernet training and maintaining a diverse set of solutions during the search. They employed the multi-objective evolutionary algorithm NSGA-II~\cite{nsga-ii}  to maximize both novelty score and architecture fitness. This approach aims to balance between exploration and the pursuit of high-performing architectures. However, they did not consider the complexity of architectures despite employing multi-objective optimization. Both of these studies calculate the novelty score based on differences in architectural designs, but  overlook information about performance and complexity metrics. Moreover, they still require training the supernets to predict the performance of candidate architectures, which incurs significant computational costs. 

\section{Methodology}\label{sec:method}

In contrast to previous studies that primarily focus on architectural novelty based on design attributes in Section~\ref{sec:ns}, our method proposes a metric-based novelty score that utilizes training-free performance and complexity metrics for multi-objective NAS. 
We argue that diversity of obtained architectures in terms of performance and complexity is more beneficial for decision makers than diversity in terms of network design.
In this section, we discuss in detail how we calculate the novelty score and enhance the exploration of new architectures while still maintaining the quality of the approximation front.

\subsection{Pareto Dominance-based Novelty Search}



\begin{algorithm}
\caption{MTF-PDNS}
\label{alg:MTF-PDNS}
\begin{algorithmic}[1]
\STATE Initialize a population $P = \{\mathbf{x}_1, \mathbf{x}_2, \ldots, \mathbf{x}_N\}$ of $N$ architectures
\STATE Initialize an empty elitist archive $\mathcal{A}$
\FOR{each architecture $\mathbf{x}_i$ in $P$}
\STATE Evaluate $\mathbf{x}_i$ using training-free and complexity metrics
\STATE Update $\mathcal{A}$ based on the evaluation of $\mathbf{x}_i$
\ENDFOR
\WHILE{termination criteria not met}
\STATE Apply crossover and mutation on $P$ to create offspring architectures $O = \{\mathbf{y}_1, \mathbf{y}_2, \ldots, \mathbf{y}_N\}$

\FOR{each architecture $\mathbf{y}_i$ in $O$}
\STATE Evaluate $\mathbf{y}_i$ using training-free and complexity metrics
\STATE Update $\mathcal{A}$ based on the evaluation of $\mathbf{y}_i$
\ENDFOR
\FOR{each architecture $\mathbf{x}$ in $P \cup O$}
\STATE Compute novelty score $\eta(\mathbf{x})$
\ENDFOR
\STATE Select architectures from $P \cup O$ based on $\eta(\mathbf{x})$
\ENDWHILE
\RETURN Elitist archive $\mathcal{A}$ as approximation front
\end{algorithmic}
\end{algorithm}

Our method is illustrated in Algorithm~\ref{alg:MTF-PDNS}. At the start of the algorithm, we randomly initialize a population $P = \{\mathbf{x}_1, \mathbf{x}_2, \ldots, \mathbf{x}_N\}$ of $N$ architectures for the genetic algorithm (line 1). Each architecture is then assessed using three training-free metrics: \texttt{synflow}, \texttt{jacov} and \texttt{snip}, along with a complexity metric (e.g., FLOPs or the number of parameters). An elitist archive $\mathcal{A}$ is created (line 2) and is considered for update each time a candidate architecture from $P$ is evaluated (lines 3-6). 

An elitist archive $\mathcal{A}$~\cite{coello_elitist_archive} is a subset of architectures that only contains non-dominated architectures from the entire architecture set explored so far. It is often incorporated in multi-objective optimization algorithms to keep track of the best solutions found during the search process. Every time a new architecture $\mathbf{y}$ is generated in the search process, we evaluate whether this new architecture $\mathbf{y}$ is dominated by any existing architecture $\mathbf{x}$ in the elitist archive $\mathcal{A}$. If $\mathbf{y}$ is not dominated by any architecture in $\mathcal{A}$, it is then added to $\mathcal{A}$. Following this addition, architectures $\mathbf{x}$ that are dominated by the newly added architecture $\mathbf{y}$ are removed from $\mathcal{A}$. The procedure of updating elitist archive $\mathcal{A}$ is detailed in the supplementary material. In our method, the elitist archive $\mathcal{A}$ serves as a memory of the search process, storing non-dominated architectures and serving as a component to compute the novelty score. It is updated dynamically after each generation and evaluation of new architectures, ensuring that the novelty scores are always relative to the current state of the search process. This dynamic updating mechanism intensifies the exploratory pressure toward novel regions within the descriptor space~\cite{ns_abonding_obj}.

In each generation, we apply crossover and mutation operations on parent architectures $P$ to create a set of offspring architectures $O$ (line 8). Similar to the initialization step, these offspring architectures are assessed using three training-free metrics: \texttt{synflow}, \texttt{jacov} and \texttt{snip}, along with a complexity metric (e.g., FLOPs or the number of parameters). The elitist archive $\mathcal{A}$ is also updated based on these metrics of offspring architectures (line 9-12).

The novelty score $\eta(\mathbf{x})$ for each architecture in both the parent set $P$ and offspring set $O$ is calculated as the mean Euclidean distance from the descriptor set of these architectures to the ones existing in the elitist archive $\mathcal{A}$ (line 13-15). The descriptor set $\phi(\mathbf{x})$ of architecture $\mathbf{x}$ includes three training-free metrics (\texttt{synflow}, \texttt{jacov}, \texttt{snip}) and a complexity metric evaluated in previous steps. It is important that these metrics should be normalized prior to the calculations of novelty scores. This normalization is crucial because it ensures that all metrics have equal weights, preventing the dominance of certain metrics due to scale differences. Without normalization, metrics with larger numerical ranges could dominate the novelty score, leading to biased search results. 
Following the computation of the novelty scores, a selection process is carried out to retain architectures from both parent set $P$ and offspring set $O$ that exhibit higher novelty scores (line 16). 
At the end of the algorithm, the final elitist archive $\mathcal{A}$ is returned as the approximation set, serving as the result of our method (line 18). Figure~\ref{fig:PDNS} illustrates our PDNS framework.

\begin{table*}[ht!]
    \footnotesize
    \centering
    \caption{Summary results on NAS-Bench-101 and NAS-Bench-1Shot1. Underline results indicate which method is significantly better when using the same performance objective (p-value < 0.01)}
    \begin{tabular}{lcccccc}\toprule
    & \multicolumn{3}{c}{\textbf{NAS-Bench-101}} & \multicolumn{3}{c}{\textbf{NAS-Bench-1Shot1}} \\
    \multicolumn{1}{c}{Methods}  & $\text{IGD}^{+}\downarrow$ &  Hypervolume $\uparrow$ & \begin{tabular}{@{}c@{}}Search cost \\ (hours) $\downarrow$\end{tabular} & $\text{IGD}^{+}\downarrow$ &  Hypervolume $\uparrow$ & \begin{tabular}{@{}c@{}}Search cost \\ (hours) $\downarrow$\end{tabular} \\
    \cmidrule(lr){1-1}\cmidrule(lr){2-4}\cmidrule(lr){5-7}
   MOENAS-\texttt{synflow} & $0.0160\pm{\scriptstyle0.0055}^-$ &  $1.0152\pm{\scriptstyle0.0074}^-$ & $0.44$ & $0.0370\pm{\scriptstyle0.0046}^-$ & $0.9989\pm{\scriptstyle0.0154}^-$ & $0.52$ \\
   PDNS-\texttt{synflow} & $\underline{0.0126\pm{\scriptstyle0.0027}^-}$ & $\underline{1.0223\pm{\scriptstyle0.0036^-}}$ & $1.45$ & $0.0364\pm{\scriptstyle0.0040}^-$ & $\underline{1.0102\pm{\scriptstyle0.0095}^-}$ & $1.28$ \\
   \midrule
   MOENAS-\texttt{jacov} &  $0.0339\pm{\scriptstyle0.0082}^-$ & $0.9843\pm{\scriptstyle0.0131}^-$ & $1.11$ & $0.0355\pm{\scriptstyle0.0054}^-$ & $0.9806\pm{\scriptstyle0.0061}^-$ & $0.91$ \\
   PDNS-\texttt{jacov} & $0.0312\pm{\scriptstyle0.0072}^-$ & $0.9901\pm{\scriptstyle0.0113}^-$ & $2.36$  &  $0.0320\pm{\scriptstyle0.0084}^-$ & $0.9865\pm{\scriptstyle0.0112}^-$ & $2.27$\\
   \midrule
   MOENAS-\texttt{snip} & $0.0648\pm{\scriptstyle0.0149}^-$ & $0.9544\pm{\scriptstyle0.0200}^-$ & $0.59$ & $0.0723\pm{\scriptstyle0.0065}^-$ & $0.9539\pm{\scriptstyle0.0138}^-$ & $0.59$\\
   PDNS-\texttt{snip} & $\underline{0.0500\pm{\scriptstyle0.0146}^-}$ & $\underline{0.9691\pm{\scriptstyle0.0173}^-}$ & $1.94$ & $0.0740\pm{\scriptstyle0.0072}^-$ & $0.9507\pm{\scriptstyle0.0136}^-$ & $1.57$\\
   \midrule
   MOENAS-\texttt{valacc} & $0.0052\pm{\scriptstyle0.0017}^-$ & $1.0279\pm{\scriptstyle0.0030}^-$ & $38.53$ & $0.0042\pm{\scriptstyle0.0012}^-$ & $1.0276\pm{\scriptstyle0.0032}^-$ & $30.66$\\
   PDNS-\texttt{valacc} & $\underline{\mathbf{0.0040\pm{\scriptstyle0.0006}^+}}$ & $\underline{\mathbf{1.0306\pm{\scriptstyle0.0017}^\approx}}$ & $104.44$ & $\underline{\mathbf{0.0033\pm{\scriptstyle0.0007}^+}}$ & $\underline{\mathbf{1.0299\pm{\scriptstyle0.0023}^\approx}}$ & $90.75$\\
   \midrule
   MTF-MOENAS & $0.0126\pm{\scriptstyle0.0034}^-$ & $1.0261\pm{\scriptstyle0.0045}^-$ & $0.94$ & $0.0123\pm{\scriptstyle0.0045}^-$ & $1.0264\pm{\scriptstyle0.0034}^-$ & $0.83$\\
   MTF-PDNS & $\underline{0.0084\pm{\scriptstyle0.0014}^-}$ & $\underline{\mathbf{1.0297\pm{\scriptstyle0.0015}^\approx}}$ & $2.02$ & $0.0100\pm{\scriptstyle0.0021}^-$ & $\underline{\mathbf{1.0295\pm{\scriptstyle0.0015}^\approx}}$ & $1.56$ \\
   \bottomrule
   \multicolumn{7}{l}{\scriptsize{$+$ \textbf{Denotes a method that delivers significantly better performance (p-value < 0.01).}}}\\
     \multicolumn{7}{l}{\scriptsize{$\approx$ Denotes a method that achieves performance comparable to the best-performing method.}}\\
     \multicolumn{7}{l}{\scriptsize{$-$ Denotes a method that delivers significantly worse performance (p-value < 0.01).}}\\
   
    \end{tabular}
        \label{tab:results_101_vs_1shot1}
    \end{table*}
    
\subsection{Novelty Score}
We compute the novelty score of an architecture $\mathbf{x}$ as:

\begin{equation}
    \eta(\mathbf{x}) = \frac{1}{|\mathcal{A}|} \sum_{i=1}^{|\mathcal{A}|} \|\phi(\mathbf{x}) - \phi(\mathbf{a}_i)\|,
\end{equation}
where $\|\phi(\mathbf{x}) - \phi(\mathbf{a}_i)\|$ represents the Euclidean distance which quantifies the difference between the descriptor set $\phi(\mathbf{x})$ of architecture $\mathbf{x}$ and the descriptor set $\phi(\mathbf{a}_i)$ of architecture $\mathbf{a}_j$ in the elitist archive $\mathcal{A}$. As mentioned above, the descriptor set of an architecture includes four values: \texttt{synflow}, \texttt{jacov} and \texttt{snip}, along with a complexity metric (e.g., FLOPs or the number of parameters). As the novelty score is a measure of how different an architecture $\mathbf{x}$ is from those $\mathbf{a}_j$ in the elitist archive $\mathcal{A}$, this encourages the exploration of a wide range of architectures, thereby reducing the risk of converging prematurely to local optima. However, such a naive approach may inadvertently assign high novelty scores to architectures that perform poorly at both performance and complexity metrics in the descriptor space. This could result in the search process expanding substantial effort exploring regions that are irrelevant to the task objectives~\cite{ns_emperical_study}.

To address this issue, we propose the use of a sign function that adjusts the novelty scores based on whether the architecture is included in the elitist archive $\mathcal{A}$. 
Architectures that belong to the elitist archive $\mathcal{A}$ can be assigned a positive novelty score, indicating that a larger score corresponds to a higher novelty. However, for architectures not included in the updated elitist archive $\mathcal{A}$, which typically correspond to poor-performing and dominated architectures, we aim to limit their survival to the next generation due to their potentially detrimental effect on the search process. Hence, these architectures are assigned a negative novelty score. This implies that a larger absolute value of these negative novelty scores indicates poorer performance, effectively discouraging the algorithm from retaining these architectures in subsequent generations. When running the algorithm, we aim to maximize this novelty score. The adjusted novelty score can be described as follows:

\begin{equation}
\eta(\mathbf{x}) =
\begin{cases}
\frac{1}{|\mathcal{A}|} \sum_{i=1}^{|\mathcal{A}|} \|\phi(\mathbf{x}) - \phi(\mathbf{a}_i)\|, & \text{if } \mathbf{x} \in \mathcal{A} \\
-\frac{1}{|\mathcal{A}|} \sum_{i=1}^{|\mathcal{A}|} \|\phi(\mathbf{x}) - \phi(\mathbf{a}_i)\|, & \text{if } \mathbf{x} \notin \mathcal{A}
\end{cases}
\end{equation}

By including both training-free performance metrics and complexity metrics in the descriptor space $\phi(\mathbf{x})$, the novelty score takes into account multiple conflicting metrics. The novelty score is also calculated based on the current elitist archive $\mathcal{A}$, which means it adapts over time as the archive evolves. This allows the algorithm to adjust its notion of \emph{novelty} based on the architectures it has seen so far, enabling a more dynamic and responsive search process. Moreover, the integration of novelty and fitness can encourage the discovery of high-quality architectures, while still preserving the diversity of architectures in the search process~\cite{ns_emperical_study}. Notably, our proposed novelty score also prioritizes assigning higher novelty scores to non-dominated architectures that appear within regions of low architectural density in our elitist archive. It promotes diversity in the population by favoring solutions in less crowded niches. In other words, this strategy encourages the algorithm to explore sparser areas, similar to NSGA-II's crowding distance principles~\cite{nsga-ii}.

    
\section{Experiments}\label{sec:experiment}

\subsection{Experimental settings}

We conduct our experiments on the following NAS benchmarks: NAS-Bench-101~\cite{nas_bench_101}, which offers a comprehensive evaluation of 423,624 architectures on CIFAR-10; NAS-Bench-1Shot1~\cite{nas_bench_1shot1}, an extension of NAS-Bench-101 that includes around 363,648 architectures; and NAS-Bench-201~\cite{nas_bench_201}, comprising 15,625 architectures assessed across multiple datasets, including CIFAR-10, CIFAR-100, and ImageNet16-120. These NAS benchmarks provide performance information for each architecture, e.g.,  accuracy, number of parameters, FLOPs, training time, etc. To assess the quality of the obtained approximation fronts, we employ the two quality indicators: Inverted Generational Distance Plus (IGD$^+$)~\cite{igd+} and Hypervolume~\cite{hypervolume}. Further details on NAS benchmarks and quality indicators can be found in the supplementary material.

When evaluating the training-free metrics for architectures in these NAS benchmarks, we utilize the pre-computed scores and the recorded computing times of the training-free metrics from \cite{zero_cost} and \cite{nas_bench_suite_zero}. Besides our main proposed method MTF-PDNS using simultaneously \texttt{synflow}, \texttt{jacov}, and \texttt{snip}, we also experiment with other four baseline versions of PDNS, each employing a different performance metric to calculate the novelty score along with the complexity metric. Similarly, we experiment with multiple versions of the multi-objective evolutionary algorithm NSGA-II~\cite{nsga-ii} for NAS (MOENAS) as baseline methods, where each optimizes performance objectives in combination with a complexity metric.
\begin{itemize}
    \item \textbf{MOENAS-\texttt{synflow}}, \textbf{MOENAS-\texttt{jacov}}, \textbf{MOENAS-\texttt{snip}}, together with \textbf{MOENAS-\texttt{valacc}}: These methods employ \texttt{syn\-flow}, \texttt{jacov}, \texttt{snip} or validation accuracy at epoch 12 respectively, as the single performance objective alongside the complexity objective to be optimized with NSGA-II.

     \item \textbf{MTF-MOENAS}: This method uses \texttt{synflow}, \texttt{jacov}, \texttt{snip} versus complexity metric as four objectives to be optimized with NSGA-II.

     \item \textbf{PDNS-\texttt{synflow}}, \textbf{PDNS-\texttt{jacov}}, \textbf{PDNS-\texttt{snip}}, and \textbf{PDNS-\texttt{val\-acc}}: These methods use \texttt{synflow}, \texttt{jacov}, \texttt{snip} or validation accuracy at epoch 12 respectively as the single performance metric, in conjunction with a complexity metric to calculate the novelty score.

    \item \textbf{MTF-PDNS (ours):} This method uses \texttt{synflow}, \texttt{jacov}, and \texttt{snip} versus a complexity metric simultaneously as multiple criteria for computing the novelty score.
\end{itemize}

The PDNS-\texttt{synflow}/\texttt{jacov}/\texttt{snip}/\texttt{valacc} variants only optimize a single objective, which is the novelty score, using the Genetic Algorithm (GA). The complexity metrics are FLOPs for NAS-Bench-201, and the number of parameters for both NAS-Bench-101 and NAS-Bench-1Shot1. PDNS employs two-point crossover and uniform mutation with population size 20. Meanwhile, MOENAS is implemented using \texttt{pymoo} framework~\citep{pymoo} with default settings as: two-point crossover, polynomial mutation, and population size 20.  Both methods run for 150 generations when applied to larger NAS benchmarks such as NAS-Bench-101 and NAS-Bench-1Shot1. 
For the smaller benchmark NAS-Bench-201, the number of generations is 50. 
Both MOENAS and PDNS utilize an elitist archive to maintain a set of non-dominated architectures obtained so far during the search. While the elitist archive in MOENAS serves solely to preserve the approximation front, the elitist archive in PDNS is also used for novelty score calculations. 
All experiments are conducted over 30 runs on a single NVIDIA GeForce RTX 3050 Ti laptop GPU.
Source codes are available at: \url{https://github.com/ELO-Lab/PDNS}.

\subsection{Qualify of Final Approximation Front}
\begin{table*}[ht!]
    \footnotesize
    \centering
     \setlength{\tabcolsep}{0.5em}
    \caption{Summary results on NAS-Bench-201. Underline results indicate which method is significantly better when using the same performance objective (p-value < 0.01)}
    \begin{tabular}{lccccccccc}\toprule
    & \multicolumn{3}{c}{\textbf{CIFAR-10}} & \multicolumn{3}{c}{\textbf{CIFAR-100}} & \multicolumn{3}{c}{\textbf{ImageNet16-120}} \\
    \multicolumn{1}{c}{Methods}  & $\text{IGD}^{+}\downarrow$ &  Hypervolume $\uparrow$ & \begin{tabular}{@{}c@{}}Search cost \\ (hours) $\downarrow$\end{tabular} &  $\text{IGD}^{+}\downarrow$ & Hypervolume $\uparrow$ & \begin{tabular}{@{}c@{}}Search cost \\ (hours) $\downarrow$\end{tabular} &  $\text{IGD}^{+}\downarrow$ & Hypervolume $\uparrow$ & \begin{tabular}{@{}c@{}}Search cost \\ (hours) $\downarrow$\end{tabular} \\
    \cmidrule(lr){1-1}\cmidrule(lr){2-4}\cmidrule(lr){5-7}\cmidrule(lr){8-10}
   MOENAS-\texttt{synflow} & $0.0126\pm{\scriptstyle0.0026}^-$ & $1.0226\pm{\scriptstyle0.0043}^-$ & $0.13$ & $0.0178\pm{\scriptstyle0.0034}^-$ & $0.7873\pm{\scriptstyle0.0054}^-$ & $0.11$ & $0.0227\pm{\scriptstyle0.0044}^-$ & $0.5080\pm{\scriptstyle0.0063}^-$ & $0.18$ \\
   PDNS-\texttt{synflow} & $0.0140\pm{\scriptstyle0.0029}^-$ & $1.0219\pm{\scriptstyle0.0030}^-$ & $0.21$ & $0.0197\pm{\scriptstyle0.0046}^-$ & $0.7836\pm{\scriptstyle0.0057}^-$ & $0.18$ & $0.0226\pm{\scriptstyle0.0056}^-$ & $0.5071\pm{\scriptstyle0.0078}^-$ &  $0.31$\\
   \midrule
   MOENAS-\texttt{jacov} & $0.0220\pm{\scriptstyle0.0098^-}$ & $1.0089\pm{\scriptstyle0.0135}^-$ & $0.14$ & $\underline{0.0332\pm{\scriptstyle0.0103}^-}$ & $\underline{0.7620\pm{\scriptstyle0.0153}^-}$ & $0.15$ & $0.0428\pm{\scriptstyle0.0171}^-$ & $0.4673\pm{\scriptstyle0.0244}^-$ & $0.19$\\
   PDNS-\texttt{jacov} & $0.0251\pm{\scriptstyle0.0097}^-$ & $1.0044\pm{\scriptstyle0.0130}^-$ & $0.23$ & $0.0501\pm{\scriptstyle0.0191}^-$ & $0.7384\pm{\scriptstyle0.0243}^-$ & $0.23$ & $0.0470\pm{\scriptstyle0.0162}^-$ & $0.4612\pm{\scriptstyle0.0233}^-$ & $0.29$ \\
   \midrule
   MOENAS-\texttt{snip} & $0.0569\pm{\scriptstyle0.0126}^-$ & $0.9660\pm{\scriptstyle0.0164}^-$ & $0.16$ & $0.1151\pm{\scriptstyle0.0221}^-$ & $0.6635\pm{\scriptstyle0.0279}^-$ & $0.16$ & $0.1541\pm{\scriptstyle0.0332}^-$ &  $0.3424\pm{\scriptstyle0.0348}^-$ & $0.18$\\
   PDNS-\texttt{snip} & $0.0630\pm{\scriptstyle0.0073}^-$ & $0.9575\pm{\scriptstyle0.0092}^-$ & $0.23$ & $0.1190\pm{\scriptstyle0.0184}^-$ & $0.6568\pm{\scriptstyle0.0226}^-$ & $0.24$ & $0.1653\pm{\scriptstyle0.0209}^-$ & $0.3308\pm{\scriptstyle0.0213}^-$ & $0.29$\\
   \midrule
   MOENAS-\texttt{valacc} & $\mathbf{0.0063\pm{\scriptstyle0.0016}^\approx}$ & $1.0286\pm{\scriptstyle0.0016}^-$ & $21.62$ & $0.0198\pm{\scriptstyle0.0091}^-$ & $0.7837\pm{\scriptstyle0.0083}^-$ & $22.27$ & $0.0199\pm{\scriptstyle0.0083}^-$ & $0.5099\pm{\scriptstyle0.0070}^-$ & $69.80$ \\
   PDNS-\texttt{valacc} & $\mathbf{0.0063\pm{\scriptstyle0.0015}^\approx}$ & $1.0283\pm{\scriptstyle0.0017}^-$ & $35.81$ & $\underline{0.0142\pm{\scriptstyle0.0037}^-}$ & $0.7875\pm{\scriptstyle0.0061}^-$ & $38.51$ & $0.0173\pm{\scriptstyle0.0055}^-$ & $0.5108\pm{\scriptstyle0.0057}^-$ & $120.06$\\
   \midrule
   MTF-MOENAS & $0.0080\pm{\scriptstyle0.0020}^-$ & $1.0300\pm{\scriptstyle0.0019}^-$ & $0.20$ & $0.0134\pm{\scriptstyle0.0032}^-$ & $\mathbf{0.7939\pm{\scriptstyle0.0047}^\approx}$ & $0.20$ & $0.0163\pm{\scriptstyle0.0046}^-$ & $0.5141\pm{\scriptstyle0.0053}^-$ & $0.23$\\
   MTF-PDNS & $\underline{\mathbf{0.0058\pm{\scriptstyle0.0015}^\approx}}$ & $\underline{\mathbf{1.0315\pm{\scriptstyle0.0011}^+}}$ & $0.28$ & $\underline{\mathbf{0.0108\pm{\scriptstyle0.0030}^+}}$ & $\mathbf{0.7965\pm{\scriptstyle0.0031}^\approx}$ & $0.28$ & $\underline{\mathbf{0.0134\pm{\scriptstyle0.0033}^+}}$ & $\underline{\mathbf{0.5180\pm{\scriptstyle0.0037}^+}}$ & $0.32$\\
   \bottomrule   
    \multicolumn{10}{l}{$+$ \textbf{Denotes a method that delivers significantly better performance (p-value < 0.01).}}\\
     \multicolumn{10}{l}{$\approx$ Denotes a method that achieves performance comparable to the best-performing method.}\\
     \multicolumn{10}{l}{$-$ Denotes a method that delivers significantly worse performance (p-value < 0.01).}\\
    \end{tabular}

        \label{tab:results_201}
    \end{table*}

Our experiment results presented across various NAS benchmarks consistently demonstrate the effectiveness of our proposed MTF-PDNS approach in terms of the quality of the approximation front. 
On NAS-Bench-101 as shown in Table~\ref{tab:results_101_vs_1shot1}, PDNS variants show a substantial increase in IGD$^{+}$ and Hypervolume compared to MOENAS variants, especially those employing a single metric like \texttt{synflow}, \texttt{snip}, \texttt{valacc}. Notably, MTF-PDNS, which employs all three training-free performance metrics (\texttt{synflow}, \texttt{jacov}, \texttt{snip}) and the complexity metric (the number of parameters) exhibits comparable performance to the training-based PDNS-\texttt{valacc}. 
MTF-PDNS achieves this result in a mere 2.02 GPU hours, dramatically less than the 104.44 GPU hours required by PDNS-\texttt{valacc}. 
Besides, the accuracy results of NAS-Bench-101 can be found in the supplementary material.

On NAS-Bench-1Shot1, the advantage of PDNS over MOENAS was not as striking when using a single training-free metric as shown in Table~\ref{tab:results_101_vs_1shot1}. However, we observed substantial improvements in both IGD$^+$ and Hypervolume when utilizing \texttt{valacc} or combining all three training-free metrics in PDNS. These improvements make PDNS-\texttt{valacc} and MTF-PDNS surpass MOENAS-\texttt{valacc} and MTF-MOENAS, respectively.
The usage of a sole training-free metric in this search space may not establish a robust correlation with actual performance, potentially causing deception and bias in the exploration during the novelty search process. In other words, the metrics must be strong correlation enough to strike a balance between exploring new architectures and efficiently exploiting existing ones. 
This balance can be achieved by combining multiple training-free metrics to calculate the novelty score as in MTF-PDNS. 
By employing this mechanism, MTF-PDNS manages to achieve significantly better IGD$^+$ and Hypervolume metrics compared to MTF-MOENAS, which directly uses training-free metrics as performance objectives. 
This improvement underscores the utility of a multi-faceted assessment approach in finding high-performing architectures, highlighting the capability of MTF-PDNS to navigate the architectural design space more efficiently and effectively. Moreover, the addition of more training-free metrics requires only a few extra seconds but yields considerable benefits. Regarding the search cost, MTF-PDNS again demonstrates its efficiency in this search space, achieving competitive results with just 1.56 GPU hours, dramatically less than the 90.75 GPU hours required by PDNS-\texttt{valacc}.

\begin{figure}[h]
     \centering
     \begin{subfigure}[b]{0.49\columnwidth}
         \centering
         \includegraphics[width=\columnwidth]{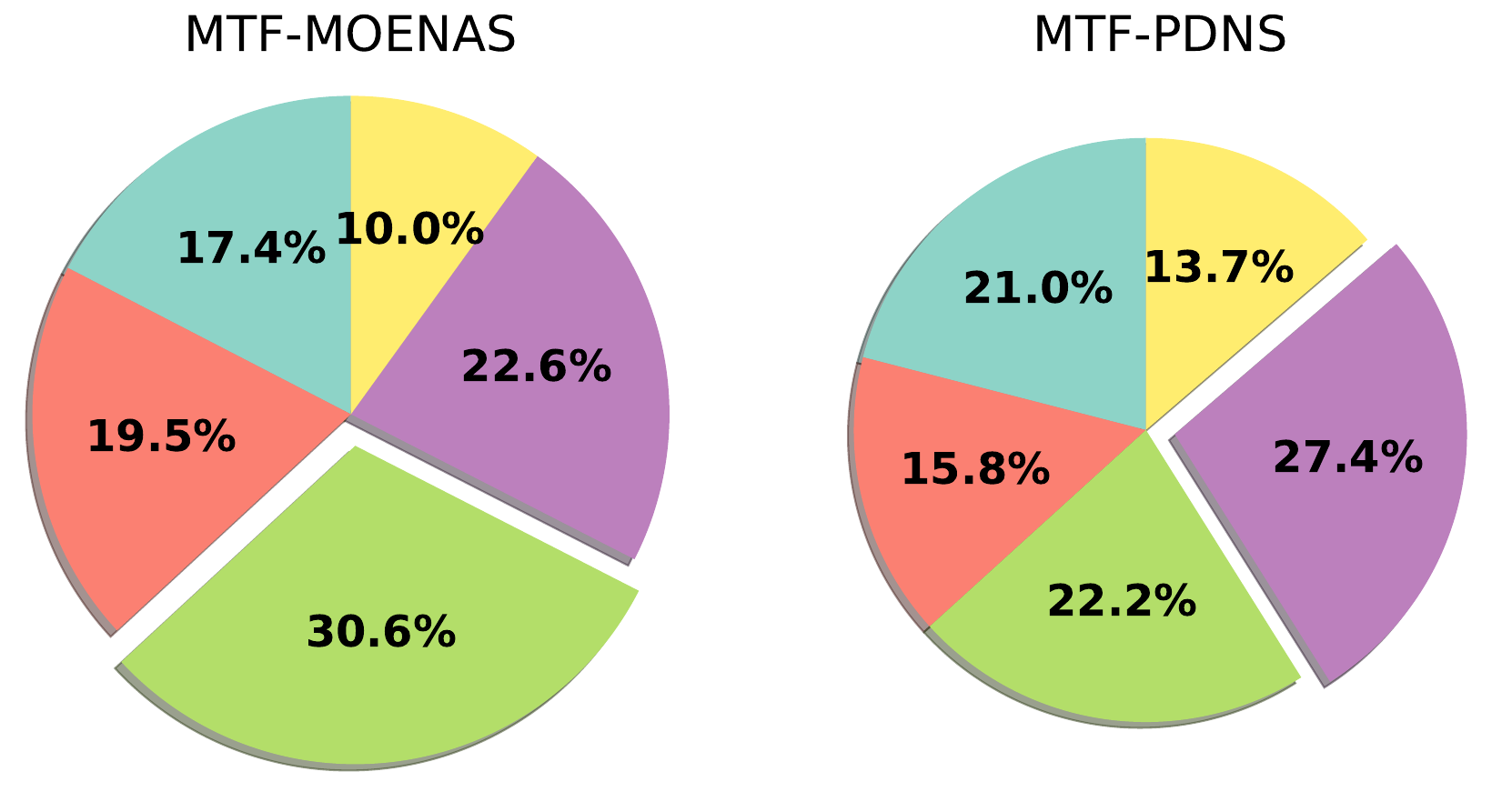}
         \caption{CIFAR-10}
         \label{fig:nb201_cifar10}
     \end{subfigure}
     \hfill
    \begin{subfigure}[b]{0.49\columnwidth}
         \centering
         \includegraphics[width=\columnwidth]{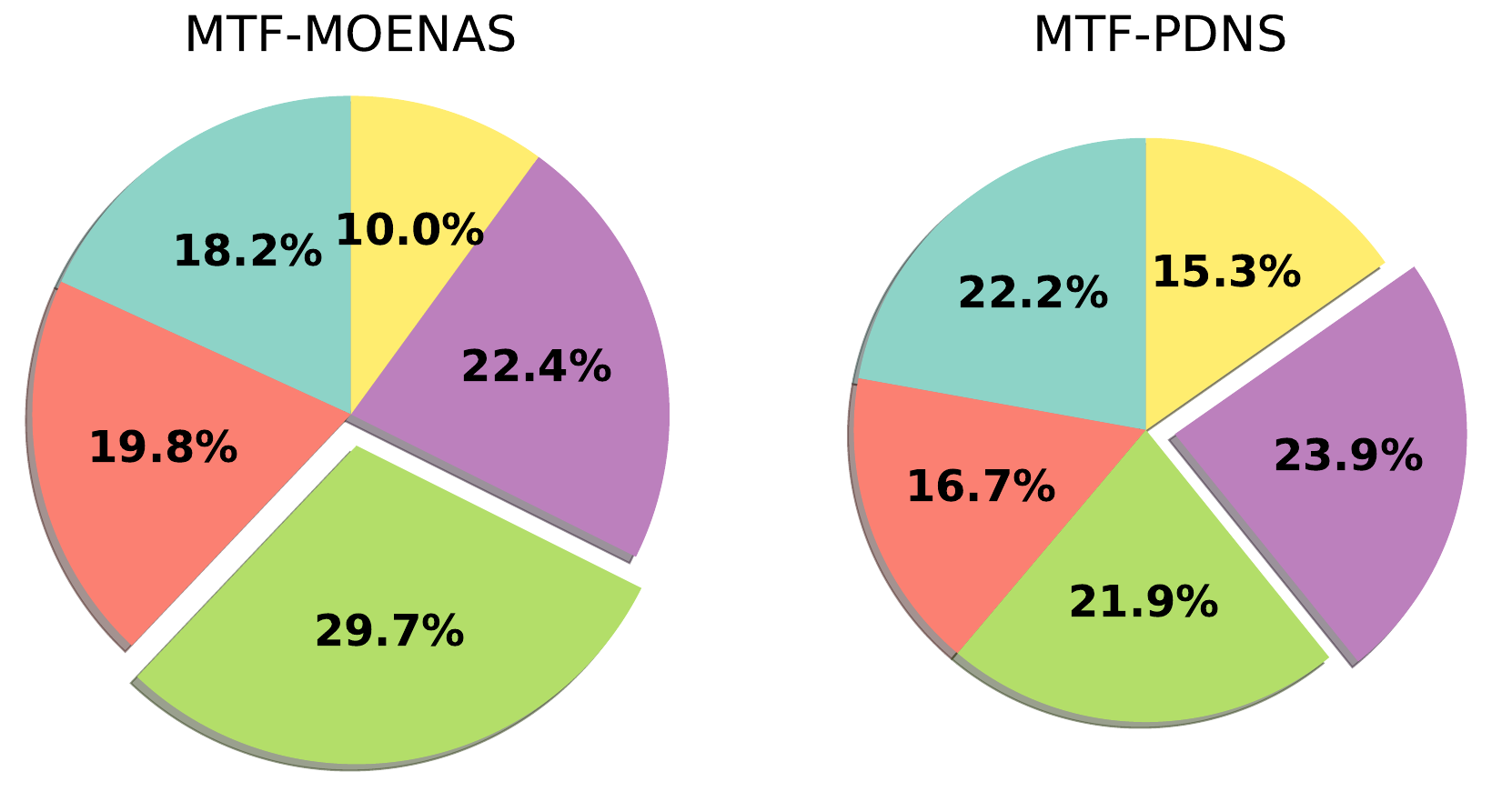}
         \caption{CIFAR-100}
         \label{fig:nb201_cifar100}
     \end{subfigure}
     \hfill
     \begin{subfigure}[b]{0.49\columnwidth}
         \centering
         \includegraphics[width=\columnwidth]{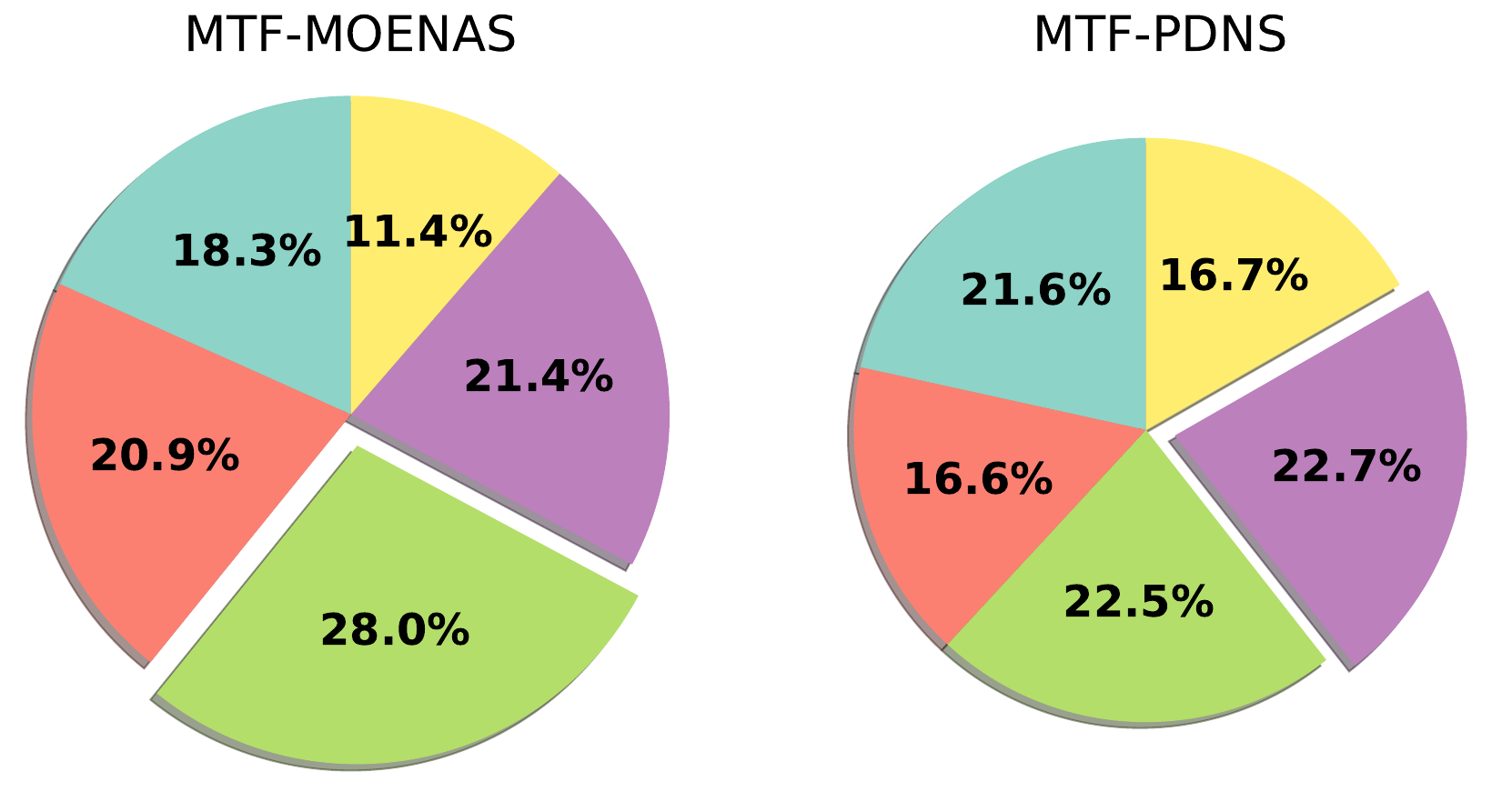}
         \caption{ImageNet16-120}
         \label{fig:nb201_ImageNet16-120}
     \end{subfigure}
     \hfill
     \includegraphics[width=\columnwidth]{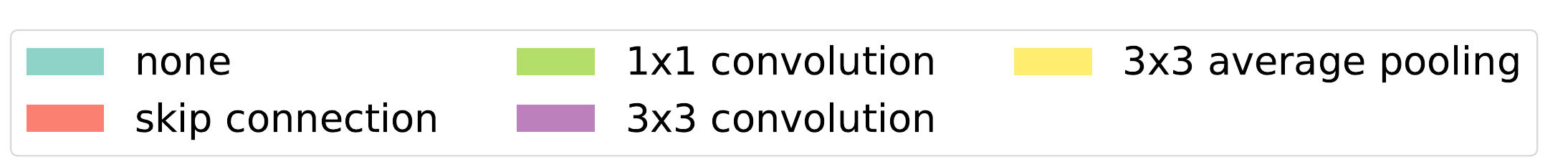}
     \caption{Distribution of operations in architectures on final approximation fronts of MTF-MOENAS and MTF-PDNS on NAS-Bench-201.}
        \label{fig:stat_ops}
\end{figure}

\begin{figure*}[ht!]
    \centering
    \includegraphics[width=0.9\textwidth]{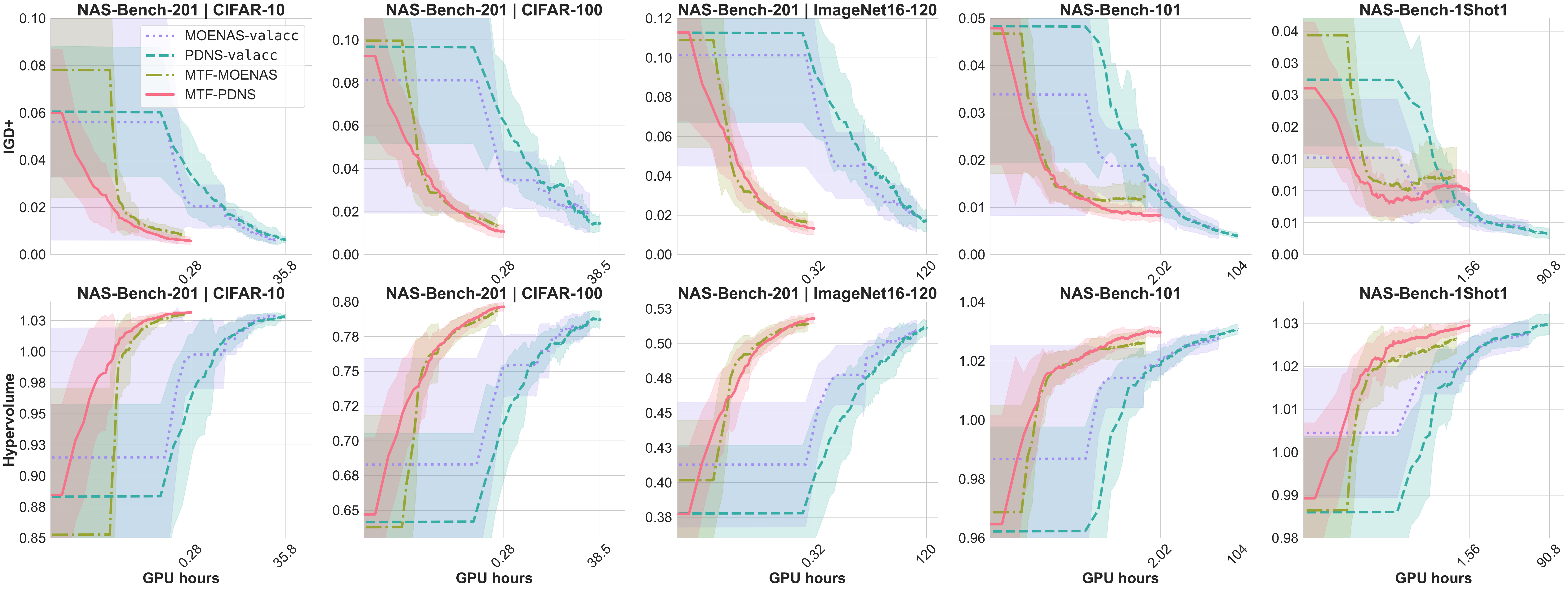}
    \caption{Comparison of IGD$^+$ (top) and Hypervolume (bottom) metrics with respect to GPU hours on NAS benchmarks.}
    \label{fig:igd_hv_results}
\end{figure*}

Table~\ref{tab:results_201} presents the results in the small search space NAS-Bench-201. We find that an increased exploration of architectures does not always lead to enhanced search performance. Despite exploring more architectures, PDNS does not always produce better results than MOENAS. 
This outcome occurs when the novelty search is guided solely by a single training-free metric, which leads to issues due to the biases of training-free metrics (discussed in Section~\ref{sec:tf_metrics}).
However, when multiple training-free metrics are incorporated, as is the case of MTF-PDNS, we see a significant performance improvement in comparison with MTF-MOENAS. 
This is similar to what we observe on NAS-Bench-101 and NAS-Bench-1Shot1. 
Using more training-free metrics as descriptors to calculate the novelty score leads to more informative evaluations and provides a more comprehensive description of an architecture's characteristics. 
MTF-PDNS also incurs a relatively low search cost compared to other methods, achieving these impressive results using about only 0.3 GPU hours across datasets. 

Figure~\ref{fig:stat_ops} illustrates the operational preferences in the design space explored by MTF-MOENAS and MTF-PDNS within NAS-Bench-201. 
Table~\ref{tab:results_201} shows that MTF-PDNS achieves higher-quality approximation fronts compared to MTF-MOENAS.
Notably, the 3x3 convolution operation predominates in the architectures on the approximation fronts obtained by MTF-PDNS, whereas the 1x1 convolution is more prevalent in those from MTF-MOENAS on all datasets of NAS-Bench-201. This observation suggests that favoring 3x3 convolution over 1x1 convolution could enhance the quality of approximation fronts in the search process.


\subsection{Efficiency and Stability}


Figure~\ref{fig:igd_hv_results} illustrates that the steeper curve of MTF-PDNS exhibits a faster convergence rate compared to other methods across different NAS benchmarks.
While using the same number of evaluations as MOENAS-\texttt{valacc} and PDNS-\texttt{valacc}, MTF-PDNS identifies high-quality approximation fronts at a quicker pace than these training-based methods. This demonstrates the efficiency of MTF-PDNS in the search process when using training-free metrics instead of using training-based metrics. In addition, MTF-PDNS consistently surpasses MTF-MOENAS across various NAS benchmarks in terms of GPU hours. MTF-PDNS consistently achieves lower IGD$^+$ and higher Hypervolume values compared to MTF-MOENAS during the entire search process. 
Note that both methods utilize training-free metrics, underscoring the effectiveness of the novelty-guided search mechanism of MTF-PDNS. 

A notable characteristic of MTF-PDNS is its stability shown in Figure~\ref{fig:igd_hv_results}. This is evidenced by its steady curve with less fluctuation in IGD$^+$ and Hypervolume. Such stability indicates that MTF-PDNS is more reliable in locating high-quality approximation fronts within just a limited number of GPU hours. The reason behind this can be attributed to the increasing number of architectures explored during the novelty search. As this number increases, the combination of multiple metrics provides a more accurate calculation of the novelty score. The combination of training-free metrics can also help to circumvent the limitations associated with individual metrics, particularly when they exhibit biases related to factors such as the cell size of architectures or the number of parameters~\cite{nas_bench_suite_zero}. 
Consequently, it results in more reliable guidance for the exploration process. 
This mechanism ensures that the search is effectively directed toward new and promising regions of the architectural space, enhancing the likelihood of discovering high-performing architectures that are different from the existing ones.

\begin{figure*}[ht!]
    \centering
    \includegraphics[width=0.9\textwidth]{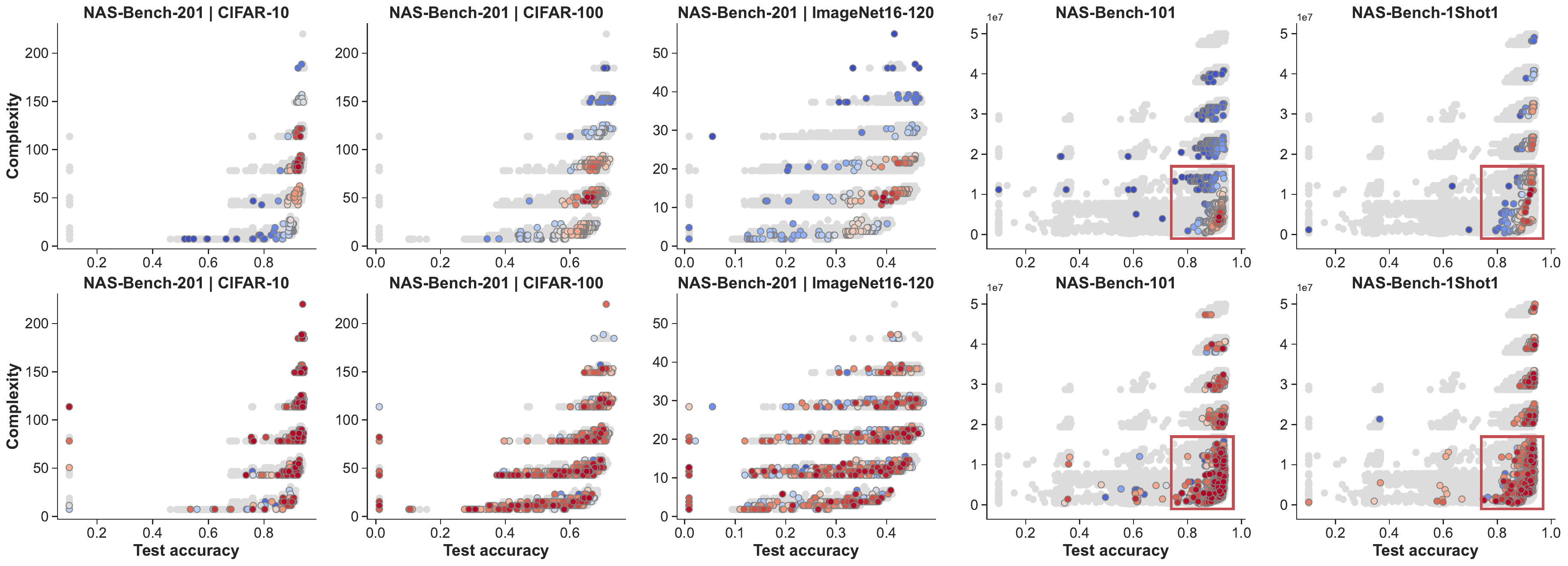}
    \includegraphics[scale=0.22]{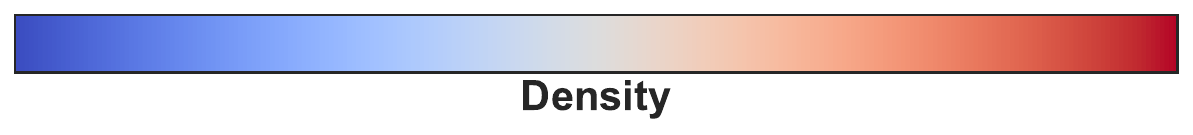}
    \caption{Visual representation of architectures explored by MTF-MOENAS (top) and MTF-PDNS (bottom) within the performance and complexity space on NAS benchmarks. Circles with no edges and blurred represent the architectures that are not explored by the corresponding search algorithm.}
    
    \label{fig:explored_archs}
\end{figure*}
\subsection{Transferability}

\begin{table}[h!]
    \footnotesize
    \centering
    \caption{Test accuracy comparison for transfer learning tasks on NAS-Bench-201}
    \begin{tabular}{lccc}
    \toprule
&  \begin{tabular}{@{}c@{}}CIFAR-10 \\ (direct)\end{tabular}  
&  \begin{tabular}{@{}c@{}}CIFAR-100 \\ (transfer)\end{tabular}  
&  \begin{tabular}{@{}c@{}}ImageNet16-120 \\ (transfer)\end{tabular} \\

\midrule
\multicolumn{4}{c}{\textbf{Evolution}} \\
REA~\cite{amoebanet} & $93.92\pm{\scriptstyle0.30}^-$  & $71.84\pm{\scriptstyle0.99}^-$ & $45.54\pm{\scriptstyle1.03}^-$  \\
G-EA~\cite{gea_gecco} & $93.99\pm{\scriptstyle0.23}^-$ & $72.36\pm{\scriptstyle0.66}^-$ & $ 46.04\pm{\scriptstyle0.67}^-$  \\ 
EvNAS~\cite{EvNAS} & $92.18\pm{\scriptstyle1.11}^-$ & $66.74\pm{\scriptstyle3.08}^-$ & $39.00\pm{\scriptstyle0.44}^-$ \\

Fast-ENAS~\cite{Fast-ENAS} & $93.74\pm{\scriptstyle0.22}^-$ & $72.00\pm{\scriptstyle0.92}^-$ & $45.74\pm{\scriptstyle0.80}^-$ \\
SF-GOMENAS~\cite{SF-GOMENAS} & $\mathbf{94.22\pm{\scriptstyle0.17}^\approx}$ & $\mathbf{72.76\pm{\scriptstyle0.63}^\approx}$ & $46.16\pm{\scriptstyle0.45}^-$  \\
PRE-NAS~\cite{PRE-NAS} & $94.04\pm{\scriptstyle0.34}^-$ & $72.02\pm{\scriptstyle1.22}^-$ & $45.34\pm{\scriptstyle1.03}^-$  \\
pEvoNAS~\cite{pEvoNAS} & $93.63\pm{\scriptstyle0.42}^-$ & $69.05\pm{\scriptstyle1.99}^-$ & $39.98\pm{\scriptstyle3.76}^-$ \\
MOENAS-\texttt{valacc}$^{\text{†}}$ & $93.86\pm{\scriptstyle0.17}^-$ & $71.14\pm{\scriptstyle0.48}^-$ & $45.11\pm{\scriptstyle1.00}^-$ \\
MTF-MOENAS$^{*\text{†}}$ & $94.14\pm{\scriptstyle0.22}^-$ & $\mathbf{72.75\pm{\scriptstyle0.68}^\approx}$ & $\mathbf{46.51\pm{\scriptstyle0.33}^\approx}$\\
\midrule
\multicolumn{4}{c}{\textbf{Novelty Search}}  \\
EN$^2$NAS~\cite{ns_nas_one_shot_sampling} & $93.36\pm{\scriptstyle0.30}^-$ & N/A & N/A \\
NEvoNAS~\cite{NEvoNAS} & $93.50\pm{\scriptstyle0.30}^-$ &  $71.51\pm{\scriptstyle0.52}^-$ & $45.30\pm{\scriptstyle0.82}^-$ \\
PDNS-\texttt{valacc}$^{\text{†}}$ & $93.81\pm{\scriptstyle0.27}^-$ & $71.00\pm{\scriptstyle0.83}^-$ & $44.93\pm{\scriptstyle1.08}^-$\\
\textbf{MTF-PDNS (ours)$^{*\text{†}}$} & $\mathbf{94.28\pm{\scriptstyle0.13}^\approx}$ & $\mathbf{73.15\pm{\scriptstyle0.53}^\approx}$ & $\mathbf{46.47\pm{\scriptstyle0.28}^\approx}$ \\

\midrule
\textbf{Optimal (in benchmark)} & $\mathbf{94.37}$  & $\mathbf{73.51} $ & $\mathbf{47.31}$ \\
\bottomrule
\multicolumn{1}{l}{* Training-Free} & \multicolumn{2}{l}{† Multi-Objective} \\
\multicolumn{4}{l}{\scriptsize{$+$ \textbf{Denotes a method that delivers significantly better performance (p-value < 0.01).}}}\\
     \multicolumn{4}{l}{\scriptsize{$\approx$ Denotes a method that achieves performance comparable to the best-performing method.}}\\
     \multicolumn{4}{l}{\scriptsize{$-$ Denotes a method that delivers significantly worse performance (p-value < 0.01).}}
\end{tabular}
    \label{tab:acc_transfer_201}
\end{table}

Table~\ref{tab:acc_transfer_201} compares the transferability of MTF-PDNS against other evolutionary and novelty search approaches on NAS-Bench-201. The experiments are initially conducted on CIFAR-10, and the architectures from the obtained approximation fronts are then transferred to CIFAR-100 and ImgageNet16-120 for evaluation. The results of transfer learning tasks are particularly crucial in real-world scenarios where the ultimate goal is to apply the discovered architectures to unseen tasks. 

In terms of accuracy, MTF-PDNS archives comparable performance with SF-GOMENAS~\cite{SF-GOMENAS} on CIFAR-10, CIFAR-100, and to MTF-MOENAS on CIFAR-100 and ImageNet16-120 and surpasses all other methods listed in Table~\ref{tab:acc_transfer_201} in identifying top-performing architectures. However, it is worth noting that SF-GOMENAS is a training-based single-objective method. While it also employs \texttt{synflow} to accelerate its solution variation, it still uses validation accuracy as the objective, requiring a training procedure that increases the search cost of SF-GOMEA. MTF-PDNS also achieves better performance compared to NEvoNAS~\cite{NEvoNAS} and EN$^2$NAS~\cite{ns_nas_one_shot_sampling}, other methods that utilize Novelty Search for NAS. Despite its high performance, MTF-PDNS utilizes completely training-free metrics in the search process, maintaining a relatively low cost compared to training-based methods. The results indicate that the best-performing architectures on the approximation fronts obtained by MTF-PDNS exhibit the ability to generalize well across different tasks. The IGD$^+$ and Hypervolume results on transfer learning task can be found in the supplementary material.

\subsection{Architectural Coverage and Prioritization}
Figure~\ref{fig:explored_archs} illustrates the architectures explored by MTF-PDNS and MTF-MOENAS on various NAS benchmarks. MTF-PDNS demonstrates its ability to cover a broader range of architectures in the search space compared to MTF-MOENAS. Notably, MTF-PDNS focuses on architectures in the high accuracy-low complexity area, as evidenced by the color density in these regions. It indicates that MTF-PDNS prioritizes not just any new architectures, but those that potentially expand the approximation front. This characteristic becomes particularly evident in larger search spaces like NAS-Bench-101 and NAS-Bench-1Shot1. The observed pattern shows that our method enhances the novelty search process while maintaining focus on exploration within valuable regions of the search space. We also have additional experiments about the efficiency in architecture exploration of MTF-PDNS and MTF-MOENAS in the supplementary material.

\section{Conclusions}\label{sec:conclusions}
In this work, we presented MTF-PDNS, the Pareto Dominance-based Novelty Search with Multiple Training-Free metrics framework for NAS, leveraging the strengths of training-free metrics and a novelty search approach.
Our method demonstrated substantial performance improvements across various NAS benchmarks, often outperforming traditional NAS methods. 
The key strength of MTF-PDNS lies in its novelty search approach, which prioritizes exploration over exploitation. 
This approach allowed MTF-PDNS to explore a broader and more diverse range of architectures, thereby avoiding premature convergence on sub-optimal solutions. 
Moreover, our training-free approach also demonstrated computational efficiency, incurring low computational costs despite the expansive exploration of the search space. 
The combination of multiple training-free and complexity metrics in MTF-PDNS provided a more holistic assessment of novelty scores of network architectures, accounting for various aspects simultaneously. 
Future works could further explore the potential of combining other training-free metrics or designing other advanced novelty search strategies.

\begin{acks}
    This research is funded by Vietnam National University HoChiMinh City (VNU-HCM) under grant number C2024-26-05.
\end{acks}
\clearpage
\balance

\bibliographystyle{ACM-Reference-Format}
\bibliography{sample-base}

\end{document}